\documentclass[conference]{IEEEtran}

\IEEEoverridecommandlockouts
\usepackage{cite}
\usepackage{amsmath,amssymb,amsfonts}
\usepackage{algorithmic}
\usepackage{graphicx}
\usepackage{textcomp}
\usepackage{xcolor}
\usepackage[a4paper, hmargin=13mm, vmargin={25mm,30mm}, headsep=4mm]{geometry}

\def\BibTeX{{\rm B\kern-.05em{\sc i\kern-.025em b}\kern-.08em
    T\kern-.1667em\lower.7ex\hbox{E}\kern-.125emX}}
\begin{document}

\title{Real-time HOG+SVM based object detection using SoC FPGA for a UHD video stream 
\thanks{The work presented in this paper was supported by the National Science Centre project no. 2016/23/D/ST6/01389 entitled ''The development of computing resources organization in latest generation of heterogeneous reconfigurable devices enabling real-time processing of UHD/4K video stream''}
}

\author{\IEEEauthorblockN{Mateusz Wasala}
\IEEEauthorblockA{\textit{Department of Automatic Control and Robotics} \\
\textit{AGH University of Science and Technology}\\
Krakow, Poland \\
wasala@agh.edu.pl} 
\and
\IEEEauthorblockN{Tomasz Kryjak}
\IEEEauthorblockA{\textit{Department of Automatic Control and Robotics} \\
\textit{AGH University of Science and Technology}\\
Krakow, Poland \\
tomasz.kryjak@agh.edu.pl} 
}

\maketitle

\begin{abstract}
Object detection is an essential component of many vision systems.
For example, pedestrian detection is used in advanced driver assistance systems (ADAS) and advanced video surveillance systems (AVSS).
Currently, most detectors use deep convolutional neural networks (e.g., the YOLO -- You Only Look Once -- family), which, however, due to their high computational complexity, are not able to process a very high-resolution video stream in real-time, especially within a limited energy budget.
In this paper we present a hardware implementation of the well-known pedestrian detector with HOG (Histogram of Oriented Gradients) feature extraction and SVM (Support Vector Machine) classification.
Our system running on AMD Xilinx Zynq UltraScale+ MPSoC (Multiprocessor System on Chip) device allows real-time processing of 4K resolution (UHD -- Ultra High Definition, 3840 x 2160 pixels) video for 60 frames per second.
The system is capable of detecting a pedestrian in a~single scale.
The results obtained confirm the high suitability of reprogrammable devices in the real-time implementation of embedded vision systems.
\end{abstract}


\section{Introduction}


Real-time object detection is crucial in robotic systems with so-called vision feedback.
Examples are advanced video surveillance systems (AVSS), advanced driver assistance systems (ADAS), or perception systems for autonomous vehicles (cars, drones).
In the discussed cases, high efficiency and low latency, i.e. the time from image acquisition to object recognition, are crucial.
Furthermore, at least in some applications, a~limited energy budget is an important factor.
Meeting these requirements at the same time is a major challenge.
High efficiency can be achieved by using good-quality cameras with high resolutions, including 4K/UHD (Ultra High Definition -- $3840 \times 2160$ pixels) and appropriate detection algorithms.
Unfortunately, these are characterised by high computational complexity, so it is often necessary to use special platforms on which hardware acceleration is possible.
Examples are heterogeneous embedded GPUs (Graphical Processing Units) and SoC FPGAs (System on Chip Field Programmable Gate Arrays).
In this paper, we have used the latter solution, as it performs better in the considered system.

Object detection can be implemented using ``classical'' methods and those based on deep convolutional neural networks (e.g., the YOLO family -- You Only Look Once \cite{Yolo2021, Yolo2018}).
In the proposed system, we have decided to use the HOG+SVM (Histogram of Oriented Gradients and Support Vector Machine) method proposed in 2006 by Dalal and Triggs \cite{Dalal}.
It has been the basis of many detection systems over the years, especially pedestrian detection.
It should be noted that, to the best of our knowledge, the use of DCNNs for real-time 4K resolution stream processing, especially in embedded scenarios, is currently not feasible.
Inputs to the network are usually scaled to relatively small sizes, e.g. $608 \times 608$ px.
For example, in the work \cite{Yolo4K} a two-stage pedestrian detection system was proposed using a YOLOv2 neural network.
Two square images of $2160 \times 2160$ pixels were cropped from the 4K input image and then scaled to the required dimensions of the network input, i.e. $608 \times 608$ pixels.
This hardware implementation was running on nodes of the PSC’s Bridges cluster and using GPUs.
The average performance of 3-6 fps on 4K video and 2 fps on 8K video was achieved.
Hence, to implement 4K video stream analysis without any rescaling, we have used the well-known ``classic'' HOG+SVM algorithm.


The main contributions of our work can be summarised as follows.
To our best knowledge, we present the first real-time HOG+SVM pedestrian detection system implemented for a~4K video stream.
This was possible due to some modification to the algorithm and the use of a modern SoC FPGA device.
Processing a 4K video allows to detected small (far from the camera) pedestrians, which is important in ADAS.


The remainder of this paper is organised as follows.
Section \ref{sec:hog_svm} presents basic information about the HOG and SVM algorithms.
Then, Section \ref{sec:related} discusses the most important related work on the hardware implementation in FPGA of the algorithms considered.
The proposed system is described in Section \ref{sec:proposed}, while the results obtained are summarised in Section \ref{sec:results}.
The paper ends with a conclusion and future research discussion.


\section{Histograms of Oriented Gradients and Support Vector Machine}
\label{sec:hog_svm}

In this section, we present a brief overview of the HOG+SVM.
We limit our discussion to the default version described in the original paper \cite{Dalal}.

The Histogram of Oriented Gradients (HOG) is a descriptor based on information about the intensity and orientation of edges.
The corresponding arrangement of edges in the detection window describes the silhouette of pedestrians (and also other objects) relatively well.
Due to the simple operations included in the descriptor, the consequently low computational complexity and the possibility to decompose the individual steps into parallel operations, as well as the relatively high performance, the HOG algorithm has gained recognition in the scientific community and is the most widely used classical method for pedestrian detection.


The HOG algorithm consists of the following steps:
\begin{enumerate}
    \item  gradient calculation (magnitude and orientation),
	\item  histogram of the orientation of gradients in the cells (size $8 \times 8$ pixels) computation,
	\item  normalisation of histograms in blocks of cells ($2 \times 2$ cells, consecutive blocks overlap at 50\%) and determining the final feature vector.
\end{enumerate}
The obtained feature vector is then classified, usually using an SVM.

In the version described above, the HOG+SVM approach allows the detection of objects in a window of a given size -- typically $64 \times 128$ pixels.
Full-resolution image analysis is achieved using the sliding window technique, and detection of objects of different sizes is achieved through a multi-scale approach.



\subsection{Gradient computation}

A context operation is performed for each pixel in the image (located at coordinates $(x,y)$). 
A one-dimensional mask set consisting of vectors  $\left[ -1 \quad 0 \quad 1 \right]$ and~$\left[ -1 \quad 0 \quad 1 \right]^T$ is used. 
As a result, the horizontal $G_x (x,y)$ and vertical $G_y (x,y)$ components of the gradient are determined.
Then, the magnitude and orientation of the gradient are calculated according to the formulas:
\begin{equation}
\label{eq:hog_magnitude}
m(x,y) = \sqrt{G_x(x,y)^2 + G_y(x,y)^2}
\end{equation}
\begin{equation}\label{eq:hog_orientation}
  \theta (x,y) = \arctan \left( \frac{G_y(x,y)}{G_x (x,y)} \right)
\end{equation}

\subsection{Histogram computation}

After calculating the magnitude and orientation of the gradient, a 9-interval histogram (in the~range  $0 - 180^\circ$ (unsigned gradient)) is computed for each cell.
The magnitude values of the gradients are distributed between adjacent intervals using bilinear interpolation.
The value added to two adjacent intervals of a histogram is calculated as follows:

\begin{align}\label{eq:hog_bilinear_interpolation}
  h(\theta_{LB}) & = h(\theta_{LB}) + m(x,y) \left(\frac{\theta_{UB} - \theta(x,y)}{d_{\theta}} \right) \\
  h(\theta_{UB}) & = h(\theta_{UB}) + m(x,y) \left( \frac{\theta(x,y)-\theta_{LB}}{d_{\theta}} \right)
\end{align}
where:
$ h(\theta_k) $ --  histogram value for the interval with centre $\theta_k$,
$ x,y $ --  pixel coordinates,
$ d_{\theta} $ -- width of the histogram interval,
$ \theta_{LB} $ -- centre of the lower histogram interval,
$ \theta_{UB} $ -- centre of the upper histogram interval.


\subsection{Block Normalisation}

The gradient value inside the detection window varies over a wide range due to local variations in lighting and background contrast.
Hence, a~local normalisation of histograms is desirable.
In the first step, the values from~the histograms are combined into a 36-element feature vector $f_k$:
\begin{equation}\label{eq:hog_feature_vector}
  f_k  = \left[ H_{i,j}^k, H_{i+1,j}^k H_{i,j+1}^k H_{i+1,j+1}^k \right]
\end{equation}
where:
$ H_{i,j}^k $ -- 9-element vector with histogram values for cell $(i,j)$ in block $k$,
$ f_k $ --  36-element feature vector for block $k$.

Then, each feature vector $f_k$ is normalised according to the Eq. \eqref{eq:hog_norma}.
This is the \textit{L2-hys} norm, which consists of two steps.
First, the norm \textit{L2} is determined, then the upper threshold with $0.2$ is applied (values greater than the threshold take the threshold value), and finally the \textit{L2} normalisation is performed again.
\begin{align}\label{eq:hog_norma}
  f_k^{l2} & = \frac{f_k}{\sqrt{\left\lVert f_k \right\rVert^2_2 + \epsilon^2}}           \\
  f_k^{th} & = \max(0.2, f_k^{l2})                                                        \\
  f'_k     & = \frac{f_k^{th}}{\sqrt{\left\lVert f_k^{th} \right\rVert^2_2 + \epsilon^2}}
\end{align}
where:
$ f_k^{l2} $ -- feature vector after \textit{L2} normalisation,
$ f_k^{th} $ --  feature vector after thresholding,
$ f'_k $ --  the resulting feature vector after normalisation,
$ \left\lVert \cdot \right\rVert_2 $ -- the Euclidean norm of the vector,
$ \epsilon $ -- a~small constant (prevents the division by 0).



Finally, the feature vector $f$, built from~elements $f'_k$, has the size of $9\times 4 \times 7 \times 15 = 3780$ elements.

\subsection{Support Vector Machine}

The SVM algorithm, in the basic version, is a binary, linear classifier.
The learning process finds the parameters of a~hyperplane that is maximally distant from all multidimensional samples (feature vectors). 
The classifier is described by the formula:
\begin{equation}\label{eq:hyperSVM}
s(f) = w^{T} f + b
\end{equation}
where:
$f$ -- feature vector,
$w$ -- weight vector -- parameters of the hyperplane,
$b$ -- bias.

If $s(f) > 0$ then the analysed detection window has been classified as containing a human silhouette. 

\section{Related Work}
\label{sec:related}


Many variations of the HOG algorithm have been developed over the years. 
Some authors have focused on a more accurate description of the silhouette of the pedestrian, introducing additional information into the histogram, and increasing the accuracy of detection \cite{PiHOG, CoHOG}.
Preprocessing methods were used in the \cite{GFHOG, HOGG} works to emphasise the contours of the pedestrians and increase the efficiency of distinguishing the pedestrian from the background and other objects. 
In other works, modifications have been made to the original structure of the HOG algorithm from \cite{Dalal} to reduce computational complexity and use embedded computing platforms for parallelisation and computation acceleration.
In this short review, we focus on the latter group, as it is directly relevant to our work.



In hardware implementation, two approaches are used: window-based \cite{Bauer, Kadota, CoHOG} and cell-based \cite{Mizuno}. 
In the window-based approach, HOG features of 105 blocks are collected. 
The features are then multiplied by the SVM coefficients corresponding to one window.
On the other hand, the cell-based approach provides HOG features after normalisation for one block. 
Then the features are multiplied by SVM coefficients corresponding to 105 blocks.
The cell-based pipeline architecture is less resource consuming, hence we used it in our implementation.


In the paper \cite{Kadota}, simplifications have been applied to the complex operations involved in gradient calculation, such as the determination of the magnitude (square root) and the orientation of the gradient (trigonometric operation). 
Furthermore, division and square root were eliminated in the histogram normalisation stage.
The value of the norm has been approximated by a value of $2^n$, making it possible to use a bit shift.
In \cite{Chen}, the authors implemented approximation methods for computationally complex operations such as square root, division, inverse square root, trigonometric functions.
They were replaced by simple bit shift operations, LUT array and the fast inverse square root method.
The proposed approximations reduced resource consumption on the FPGA platform, achieving a higher operating frequency while not significantly affecting accuracy and performance compared to the original HOG algorithm.
In the work \cite{Hsiao}, a significant reduction in the information of the silhouette of pedestrians was applied.
It involved substituting the gradient magnitude and orientation with binary signed digit values and a LUT array to determine the histogram interval membership.  
A simple binarization was used to normalise the histogram. 
The proposed hardware implementation had low resource consumption, but the simplifications introduced reduced the detection rate of the algorithm on the INRIA dataset (the authors reported 93.9 \%).


In the works discussed so far, the hardware implementations were adapted to process data with a maximum resolution of Full HD ($1920 \times 1080$ px). 
In this paper, we describe a~hardware implementation of the HOG+SVM algorithm for 4K UHD resolution ($3840 \times 2160$ px) implemented on a ZCU 104 FPGA SoC platform with a Zynq UltraScale+ MPSoC chip from AMD Xilinx.




\section{The proposed  HOG + SVM implementation}
\label{sec:proposed}

\begin{figure}[!t]
\centerline{\includegraphics[width=1\linewidth]{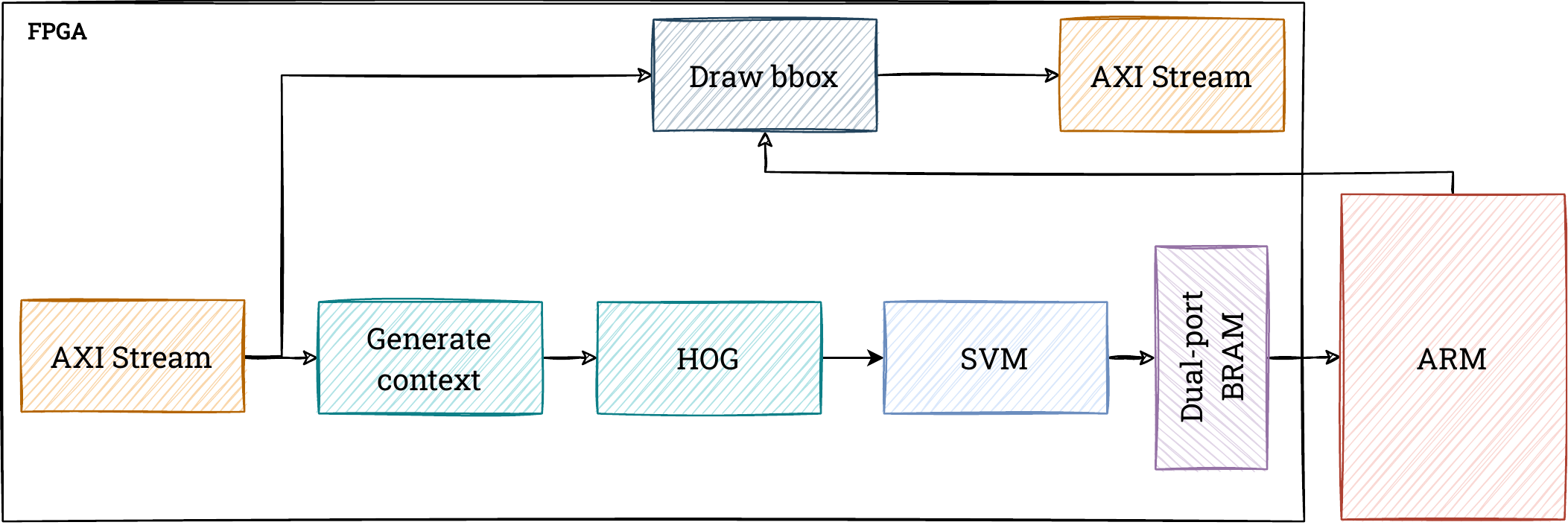}}
\caption{Overall scheme of our hardware implementation of HOG+SVM on an SoC FPGA platform.}
\label{fig:scheme:hog_scheme}
\end{figure}

Processing 4K UHD video in SoC FPGAs is a significant challenge.
The 4K signal contains 4x more data than Full HD ($1920 \times 1080$ px), which, assuming the same frequency (typically 60 fps), means the need to increase the so-called pixel clock from about 150 MHz to 600 MHz.
Using such a clock value in currently available reprogrammable devices is impossible (except for very simple logic elements).
Hence the necessity of processing data in 2 or 4 pixels per clock (ppc) vector format, which reduces the pixel clock to 300 MHz and 150 MHz respectively \cite{Kowalczyk}.
This data format implies the use of a different approach during the implementation of individual operations in the HOG+SVM algorithm.
Thus, the performance of the system described in this paper in not only the result of using a modern SoC FPGA platform, but also the appropriate design of particular computational modules.
The general scheme of our hardware implementation is shown in Figure \ref{fig:scheme:hog_scheme}.
This architecture processes the video stream in a fully-pipelined manner using a sliding window approach.
Thus, after some input latency, the detection scores for each windows are provided at the output.
The input to the module is a greyscale stream in the 4 ppc format (the AXI-Stream databus is used).
Note that in the next subsections, we present our solution in the 4 ppc format, but the same approach can be used to 2 ppc or 8 ppc.



Based on the work of \cite{Kadota, Chen}, we have decided to use approximations of particular elements of the HOG algorithm to reduce the computational complexity of the methods, that is, square root, trigonometric function, and inverse square root.
We also used fixed-point calculations.
Their precision was adjusted on the basis of preliminary experiments and is shown in Table \ref{tab:hog_svm_precision}. 



\begin{table}[!t]
	\caption{Fixed-point precision applied at different stages of the HOG+SVM algorithm.}
	\label{tab:hog_svm_precision}
	\begin{center}
		\begin{tabular}{|c|c|c|}
			\hline
			Name            & Width & Precision          \\ \hline
			Gradient magnitude & 11 & 3                            \\ \hline
			Histogram bin number & 4 & 0              \\ \hline
			Histogram value &  18 & 4              \\ \hline
			Prepare to first normalisation &  42 & 8              \\ \hline
			First inverted square root &  24 & 18              \\ \hline
			Feature vector after first normalisation &  10 & 9              \\ \hline
            Second inverted square root &  22 & 16              \\ \hline
			Final feature vector &  10 & 9              \\ \hline
			SVM coefficient &  11 & 10              \\ \hline
			SVM bias &  33 & 19              \\ \hline
			SVM prediction & 33 & 19 \\ \hline
		\end{tabular}
	\end{center}
\end{table}


\subsection{Gradient computation}

The root and square root operations are computationally intensive. 
Therefore, we have used the square root approximation (SRA) technique proposed in papers \cite{Gajski, Chen, Wang} to determine the magnitude of the gradient $m(x,y)$. 
Thus, we were able to use bit shift, which implementation in FPGAs is very simple (appropriate routing).
Equation \eqref{eq:hog_magnitude} has been replaced by the functions $max$ and $min$ according to:
\begin{equation}\label{eq:hog_mod_amplituda_SRA}
	\begin{aligned}
		m(x,y) & = \sqrt{G_{x}(x,y)^{2} + G_{y} (x,y)^{2}} \\ & \approx \max ((0.875a + 0.5b), a) \\
		a      & = \max(\left| G_{x} (x,y)\right| ,\left| G_{y} (x,y)\right| )               \\
		b      & = \min(\left| G_{x} (x,y)\right| ,\left| G_{y} (x,y)\right| )
	\end{aligned}
\end{equation}

The next step is to compute the gradient orientation according to the Eq. \eqref{eq:hog_orientation}.
In hardware implementations, the most commonly used method for calculating various trigonometric functions is the CORDIC algorithm (Coordinate Rotation DIgital Computer).
It involves the use of a series of bit-shift and addition operations.
Its major drawback is the use of multiple iterations to obtain the best possible precision.
For this reason, we have abandoned the exact calculation of the gradient angle in favour of assigning it to an appropriate interval. 
This type of approach has been proposed in the work \cite{Chen}.
Equation \eqref{eq:hog_orientation} is converted into the inequality Eq. \eqref{eq:hog_mod_orient_approx}:
\begin{equation}\label{eq:hog_mod_orient_approx}
	\begin{aligned}
		\tan\theta_{i}            & \leq & \tan\theta (x,y) & \leq  \tan\theta_{i+1}             \\
		G_{x}(x,y) \tan\theta_{i} & \leq & G_{y} (x,y)      & \leq  G_{x} (x,y) \tan\theta_{i+1}
	\end{aligned}
\end{equation}
where:
$\theta(x,y)$ -- gradient orientation,
$G_{x}(x,y),G_{y}(x,y)$ -- gradient components,
$\tan\theta_{i},\tan\theta_{i+1}$ -- the tangent values of two adjacent intervals to which the gradient will be assigned, values from the work \cite{Chen}.

The tangent values on the right and left sides of the inequality are computed offline and stored in a LUT (look-up table) -- the relevant values were used from the work \cite{Chen}.
Particular values are approximated by numbers that can be obtained by simple addition and bit-shift operations.
The proposed tangent values are only analysed for the first quadrant of the coordinate system.
The angles in the second quadrant can easily be reduced to the first when we change the sign of the horizontal component of the gradient: $ G_{x}(x,y) = -G_{x}(x,y)$.
As a consequence, the memory needed for LUT arrays is reduced.

\begin{figure}[!t]
\centerline{\includegraphics[width=.8\linewidth]{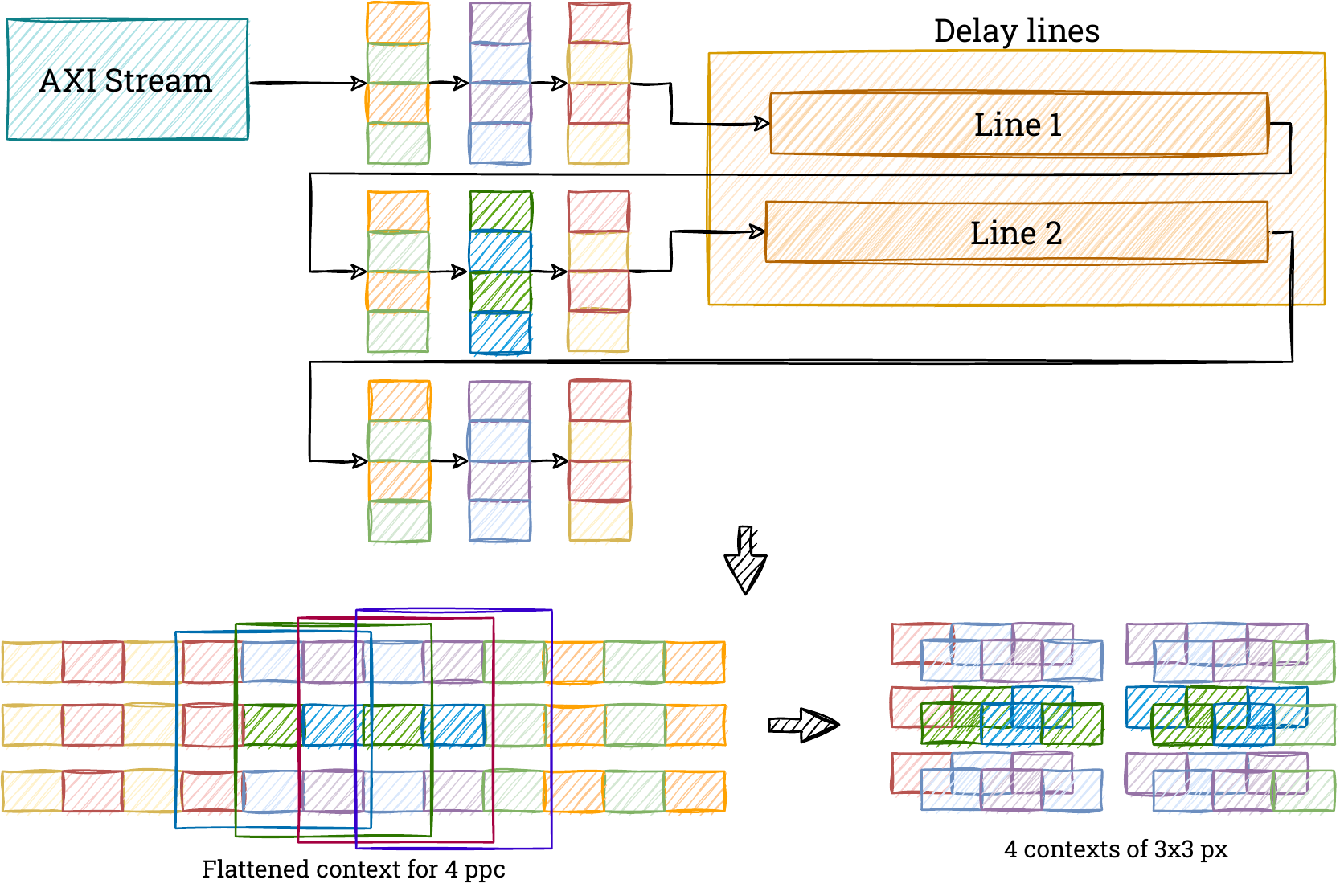}}
\caption{Context computation for data in the 4 ppc format.}
\label{fig:scheme:hog_ppc}
\end{figure}

In our design, we adapted pipelined processing to compute the magnitude and orientation of the gradient in parallel. 
The first step is to recreate a $3 \times 3$ pixel context for each pixel in the 4 pixels vector. 
For this we used registers to store the pixel values and delay lines with BRAM memory.
A context module for the 4 ppc format is presented in Figure \ref{fig:scheme:hog_ppc}.
More details on this issue are available in \cite{Kowalczyk}.


Then, for each pixel in the vector we determine the gradients $G_{x}(x,y), G_{y}(x,y)$, and calculate the magnitude and the two adjacent histogram intervals to which the magnitude value should be assigned (thus orientation).
The output of the gradient module still retains the 4 ppc format. 


\subsection{Histogram computation}


Once we have the magnitude and the intervals that define the approximate orientation of the gradient, we generate a~histogram for each cell. 
We replace the linear interpolation with a uniform distribution of values between intervals, since only the approximate orientation of the gradient is known:  

\begin{align}\label{eq:hog_mod_przedzial_approx}
	\upsilon_{UB}(x,y) & = 0.5 \: m(x,y) = m(x,y) \gg 1 \\
	\upsilon_{LB}(x,y) & = 0.5 \: m(x,y) = m(x,y) \gg 1
\end{align}



\begin{figure}[!t]
\centerline{\includegraphics[width=1\linewidth]{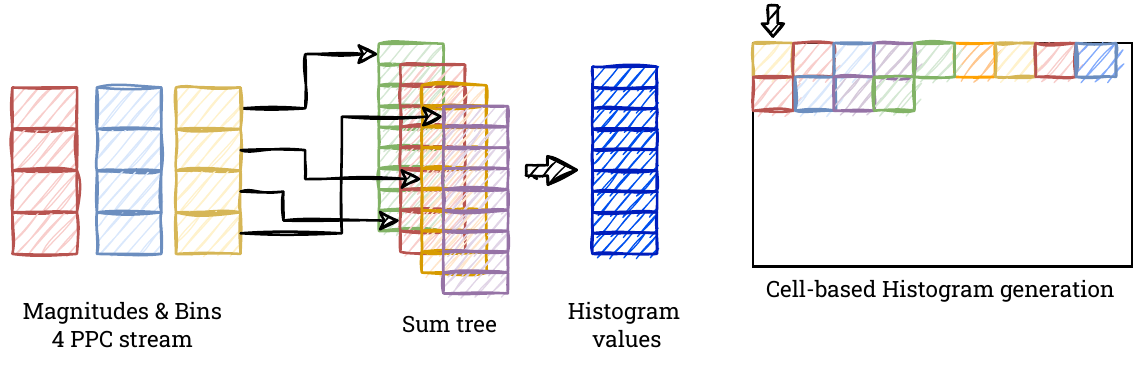}}
\caption{Hardware implementation of cell-based histogram calculation.}
\label{fig:scheme:hog_histogram}
\end{figure}

The most straightforward approach to histogram computation in FPGA is the use of BRAM (Block RAM) resources.
However, in this case, simultaneous access to multiple values from multiple addresses is not possible, and reading and writing take several clock cycles.
On the other hand, for a~4 ppc pixel vector, four accesses are needed.
So, we have decided to use registers to store the values of the histograms.
This allows us to get unlimited access to all the histogram elements. 
Each histogram is created using values from a $8 \times 8$ pixels cell. 
The 4 ppc data stream contains information about the magnitude of the gradient and the interval to which the gradient belongs. 
It should be noted that the case where all the values in one data vector need to update one histogram interval should also be considered.
We solved it using summation trees to accumulate the histogram values.
Once a histogram is generated for a cell, it is sent as a~9-element vector to the next step, which is normalisation. 
The proposed approach is shown in Figure \ref{fig:scheme:hog_histogram}.
It should be noted, the instant availability of the vector containing the histogram values makes it unnecessary to use double buffering, thus saving resources.
By the time all histograms from one line of the image are finished, the initial registers from this line will already have been prepared to accumulate data from subsequent cells (next line).

\subsection{Block Normalization}



The final step is to normalise the histograms within each block. 
In the hardware implementation, finding the inverse square root according to the Eq. \eqref{eq:hog_norma} is complex and resource consuming. 
To simplify this, we used the Newton-Rapson method and the magic number i.e. \texttt{0x5F3759DF} to determine the root value.
This comes down to a fast computation of the inverse of the square root of $\dfrac{1}{\sqrt{x}}$, proposed in \cite{fastinv}.
First, we convert the $x$ value to the IEEE754 single precision format.
Then, we perform the operations according to:
\begin{equation}\label{eq:hog_mod_norma_approx}
	\begin{aligned}
      y_{n_{IEEE754}}   & = x_{IEEE754} \gg 1 - 0x5f3759df \\
      y_{approx} 	    & = \dfrac{y_{n_{IEEE754}}(3-x y^2_{n_{IEEE754}})}{2}
  	\end{aligned}
\end{equation}
Finally, we multiply the approximated value $y_{approx}$ by the feature vector \textit{f} associated with the corresponding block:
\begin{equation}\label{eq:hog_mod_norma_approx_wektor}
  f_{norm} = f \cdot y_{approx}
\end{equation}

\begin{figure}[!t]
\centerline{\includegraphics[width=1\linewidth]{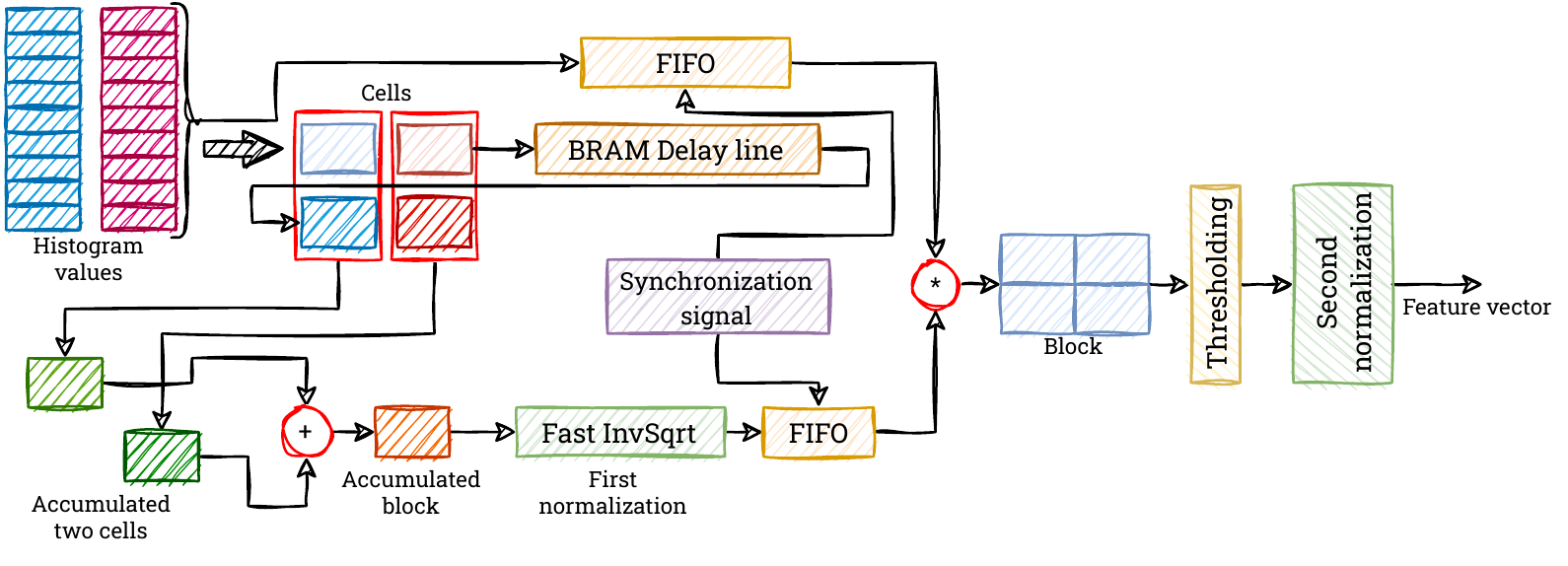}}
\caption{Hardware architecture for block histogram normalisation.}
\label{fig:scheme:hog_norm}
\end{figure}

Our approach to block normalisation, using the methods described above, is shown in Figure \ref{fig:scheme:hog_norm}.
Before determining the inverse of the square root, an appropriate preparation of the data stream is required to keep the operation pipelined.
The first step is to accumulate the histogram values for a~single cell -- calculating the squares of the histogram values in parallel and summing them using a summation tree.
The accumulated values are then delayed by one cell line. 
The next step is to sum the values from two adjacent cells from two consecutive cell lines.
To obtain the summed values from four adjacent cells (which are part of a single block), it is sufficient to add together the one-delayed partial sum from two cells and the one that has now been determined. 
Using this technique, we obtain the values for consecutive blocks that overlap in 50\% (two cells are common to two adjacent blocks).
In parallel to the operation of accumulating values for one block, the histogram values are stored in a FIFO queue. 
When the inverse of the square root is determined, the buffered histogram values are read out.
Since the blocks overlap, we decided to simultaneously normalise the cell values by the values belonging to the blocks (4*9 multiplication operations). 
Thus, the data stream at this stage contains four histograms (four cells) belonging to adjacent blocks. 
The result is a single normalised feature vector.
Performing thresholding and second normalisation is much simpler as we have access to histogram values from a~single block in a single clock cycle. 
The sum of the squares of the histogram values from one block is obtained by parallel squaring and applying the summation tree again. 
Then the second normalisation is performed.
The final result is a 36-element feature vector that is associated with one block.

\subsection{SVM classifier}



The implementation of approximation operations in the HOG algorithm resulted in changes in the feature vector.
For this reason, it was necessary to train a linear SVM classifier. 
For this we used the libraries offered by MathWorks MATLAB. 
We used the INRIA dataset and the Oxford Town Centre dataset for training.
The training dataset contained 2600 positive samples and 24300 negative samples.
We observed that using a greyscale image to determine HOG features significantly improves the performance of the SVM classifier compared to the RGB image.
In the learning process, we obtained an accuracy of 98\% (non-approximated model -- 98.5\%).
The hyperplane parameters obtained from the SVM classifier were saved to files. 


In the hardware implementation, the SVM parameters are loaded from files and stored in ROM during initialisation.
The fully pipelined SVM classification solution is based on the proposal from the work \cite{Meus}.
Due to the form of the incoming data containing the HOG feature vector (a 36-dimensional feature vector from one block is received in one clock cycle), we used a controller with four FIFO queues to reduce the number of resources that are used for multiplication operations. 
Dividing the feature vector into four 9-element vectors does not add additional latency to the system.
The feature vector is transmitted simultaneously to each element $ep_x$. 
After four consecutive clock cycles, the partial results of the multiplication are summed together and the partial result of the SVM classifier for one block of the detection frame is obtained.
Then they are passed to the element $ep_{x+1}$ located on the right.
After accumulating all the blocks in the frame, a value is obtained which is adjusted by the bias parameter.
Based on the classification score, the classifier predicts whether a~pedestrian is present in the considered frame. 
If a pedestrian is detected, the bounding box coordinates are returned, which are stored in the BRAM memory. 
The ARM processor of the SoC FPGA performs a simple filtering based on the Non-maximum Suppression method.
The filtered bounding boxes go back to BRAM memory and are displayed on the screen.

\begin{figure}[!t]
\centerline{\includegraphics[width=1\linewidth]{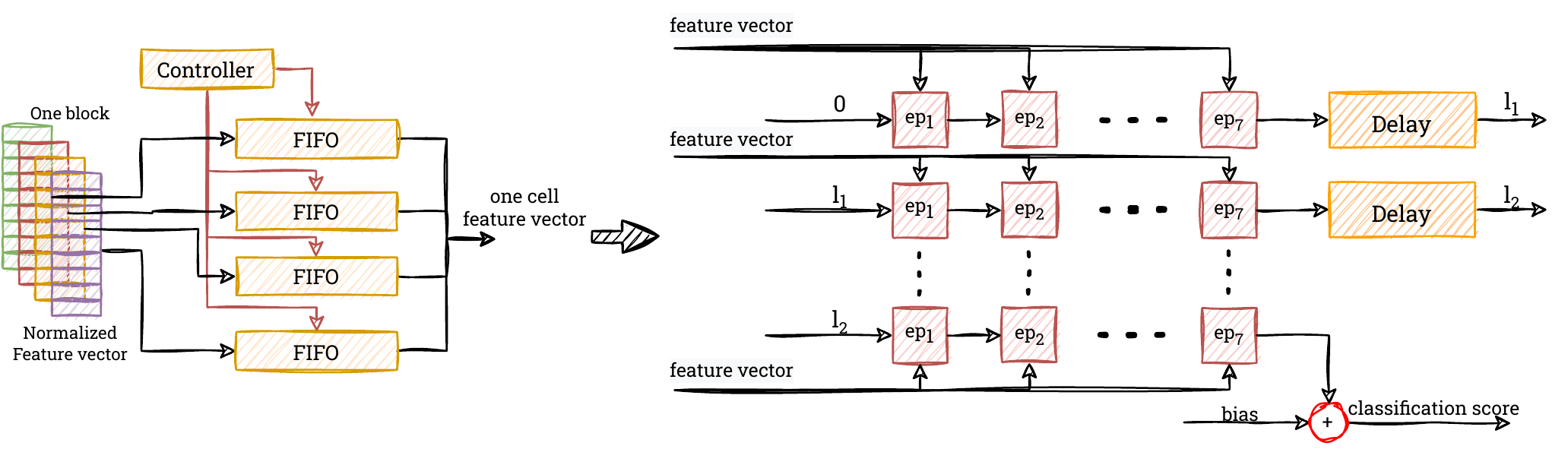}}
\caption{The proposed SVM classifier architecture.}
\label{fig:scheme:hog_svm}
\end{figure}



\begin{table*}[!ht]
\caption{Comparison of the proposed solution with other modules reported in scientific papers}
\centering
\begin{tabular}{|c|c|c|c|c|c|c|c|c|}
\hline
 & FPGA & \# of LUTs & \# of registers, FFs & BRAM & DSP blocks & FPS & Frequency (MHz) & Resolution \\ \hline
\cite{Kadota} & Altera Stratix II & 3794 & 6699 & - & 12 & 30 & 127 & 640x480 \\ \hline
\cite{Bauer} & AMD Xilinx Spartan 3 & 42435 & - & 60 & 18 & 20 & 63 & 800x600 \\ \hline
\cite{Mizuno} & Altera Cyclone IV & 34403 & 23247 & - & 68 & 30 & 76.2 & 1920x1080 \\ \hline
\cite{Hahnle} & AMD Xilinx Virtex 5 & 5188 & 5158 & - & 49 & 64 & 133 + 260 & 1920x1080 \\ \hline
\cite{Chen} & Altera Stratix II & 3467 & 172 & - & - & - & 241 & - \\ \hline
\cite{Hemmati} & AMD Xilinx Zynq 7000 & 7226 & 12462 & 12 & 26 & 60 & 125 & 1920x1080 \\ \hline
\cite{Advani} & AMD Xilinx Virtex 7 & 130329 & 102929 & 275 & 158 & 30 & 100 & 1920x1080 \\ \hline
\cite{Rettkowski} & AMD Xilinx Zedboard & 21297 & - & 0 & 4 & 40 & 82.2 & 1920x1080 \\ \hline
\cite{Saidani} & AMD Xilinx ZC702 & 27780 & - & 94 & 46 & 60 & 148 & 1920x1080 \\ \hline
\cite{Meus}& AMD Xilinx ZC702 & 24373 & 34394 & 34.5 & 116 & 60 & 74.25 & 1280x720 \\ \hline
\cite{Durre} & Altera Stratix V & 3529 & 2657 & 815 & 26 & 20 & 142 + 284 & 800x600 \\ \hline
\cite{Li} & Altera Stratix IV & 313349 & 89260 & 130 & 236 & 10000 & 80 & 512x512 \\ \hline
\textbf{Ours} & \textbf{AMD Xilinx ZCU104} & \textbf{126050} & \textbf{270817} & \textbf{38.5} & \textbf{818} & \textbf{60} & \textbf{150} & \textbf{3840x2160} \\ \hline
\end{tabular}
\label{tab:resources}
\end{table*}

\section{Results} 
\label{sec:results}






The pedestrian detection system described in Section \ref{sec:proposed} was implemented on an AMD Xilinx ZCU104 board equipped with a Zynq UltraScale+ MPSoC EV, with a quad-core ARM Cortex-A53 applications processor, dual-core Cortex-R5 real-time processor and Mali-400 MP2 graphics processing unit.
The hardware implementation was written in the SystemVerilog hardware language in AMD Xilinx Vivado 2020.2. 
Our system allows real-time processing of 4K resolution (UHD, $3840 \times 2160$ px) video for 60 frames per second and is capable of detecting a pedestrian in one scale.
The proposed architecture can be adapted to another data stream format, in general X ppc, where X is a power of 2. 
For the data vector above 8 ppc (streaming data belong to more than one cell), several histograms are generated simultaneously.
The utilisation of FPGA resources for the proposed system and comparison with other FPGA implementations is presented in Table \ref{tab:resources}. 
The design uses about half of the available resources of the AMD Xilinx ZCU104 board.
It is difficult to fairly compare resources with other implementations, as the image resolution used by us is much higher and the X ppc format imposes the need to implement some computing elements X times. 
However, it should be emphasised that our implementation in particular cases (like \# of LUTs) uses much less logical resources than implementations offering Full HD or even lower image processing.
The estimated energy consumption of the pedestrian detection system is 9.579 W.


\section{Conclusions}


In this paper, we have presented a hardware-software implementation of the HOG+SVM algorithm in a~heterogeneous SoC FPGA device.
We were able to obtain real-time processing for a $3840 \times 2160$ @ 60 fps video stream with an estimated energy consumption of approximately 9.579 W.
The proposed architecture can be adapted to another data stream format, in general X ppc, where X is a power of 2. 
We have presented a comparison with all known hardware implementations using the HOG+SVM algorithm for object detection, especially pedestrians, and our proposal is the only verified for 4K/UHD resolution.
The next step in this project is the implementation of the system for multiple scales.
We also believe that the  resource utilisation and energy consumption can be further optimised.




\vspace{12pt}

\end{document}